**Development and testing of an image transformer for explainable autonomous driving systems**

**Jiqian Dong**
Graduate Research Assistant, Center for Connected and Automated Transportation (CCAT), and Lyles School of Civil Engineering, Purdue University, West Lafayette, IN, 47907.
Email: dong282@purdue.edu
ORCID #: 0000-0002-2924-5728

**Sikai Chen\***
Postdoctoral Research Fellow, Center for Connected and Automated Transportation (CCAT), and Lyles School of Civil Engineering, Purdue University, West Lafayette, IN, 47907.
Email: chen1670@purdue.edu; and
Visiting Research Fellow, Robotics Institute, School of Computer Science, Carnegie Mellon University, Pittsburgh, PA, 15213.
Email: sikaichen@cmu.edu
ORCID #: 0000-0002-5931-5619
(Corresponding author)

**Shuya Zong**
Graduate Research Assistant, Center for Connected and Automated Transportation (CCAT), and Lyles School of Civil Engineering, Purdue University, West Lafayette, IN, 47907.
Email: szong@purdue.edu

**Tiantian Chen**
Postdoctoral Research Fellow, Department of Industrial and System Engineering, The Hong Kong Polytechnic University, Hung Hom, Kowloon, Hong Kong
Email: tt-nicole.chen@connect.polyu.hk

**Mohammad Miralinaghi**
Postdoctoral Research Fellow, Center for Connected and Automated Transportation (CCAT), and Lyles School of Civil Engineering, Purdue University, West Lafayette, IN, 47907.
Email: smiralin@purdue.edu

**Samuel Labi**
Professor, Center for Connected and Automated Transportation (CCAT), and Lyles School of Civil Engineering, Purdue University, West Lafayette, IN, 47907.
Email: labi@purdue.edu
ORCID #: 0000-0001-9830-2071

Word Count: 5,127 words + 7 tables/figures (250 words per table/figure) = 6,877 words
Submission Date: 07/31/2021
Submitted for PRESENTATION ONLY at the 2022 Annual Meeting of the Transportation Research Board

**ABSTRACT**
In the last decade, deep learning (DL) approaches have been used successfully in computer vision (CV) applications. However, DL-based CV models are generally considered to be black boxes due to their lack of interpretability. This black box behavior has exacerbated user distrust and therefore has prevented widespread deployment DLCV models in autonomous driving tasks even though some of these models exhibit superiority over human performance. For this reason, it is essential to develop explainable DL models for autonomous driving task. Explainable DL models can not only boost user trust in autonomy but also serve as a diagnostic approach to identify anydefects and weaknesses of the model during the system development phase. In this paper, we propose an explainable end-to-end autonomous driving system based on "Transformer," a state-of-the-art (SOTA) self-attention based model, to map visual features from images collected by onboard cameras to guide potential driving actions with corresponding explanations. The model achieves a soft attention over the global features of the image. The results demonstrate the efficacy of our proposed model as it exhibits superior performance (in terms of correct prediction of actions and explanations) compared to the benchmark model by a significant margin with lower computational cost.

# INTRODUCTION
## Background
Autonomous vehicles (AVs) are considered a fundamental component of next-generation transportation because it incorporates emerging technologies including AI, data analytics, machine learning, and control theory. Also, AVs have a number of prospective benefits including enhancement of safety and reliability, reduction of human labor and cost, and reduction of emissions and energy consumption (*1–3*). However, to date, the deployment of AV systems in the real world has been severely limited because several key technologies associated with perception and decision processors have still not reached a level of technological advancements where they can be applied confidently in AV systems. For example, Deep learning (DL) is the mainstream technology underlying the field of computer vision (CV) and consequentky has been applied in several vision-based AV systems (*4–9*). However, there still exists opportunities to enhance DL further. In the subsequent sub-sections of this introductory section, we discuss the two major approaches for developing deep-learning computer-vision (DLCV)-based AV systems, the image-attention based technologies used to imitate human vision, and we identify some research gaps in existing research and highlight the prospective contributions of this paper.

## End-to-end vs. pipelined systems
In developing deep-learning computer-vision (DLCV)-based AV systems, there exist two major approaches: end-to-end and pipelined. The former seeks a direct map from the raw sensor inputs (including images and 3D cloud points) to the driving actions (including straight movement, left/right turn, or slowing down (*4–9*)). The latter divides the entire system into sub-systems (including vision block (*10*, *11*) and decision-generating block (*12*, *13*)) and addresses them independently. Theoretically, end-to-end approaches are superior to pipeline approaches because the vision block can get trained to be goal-induced, meaning it is capable of paying more attention to the visual information that is necessary for the ultimate goal. However, the end-to-end approach is generally more complicated and needs relatively deeper networks and larger datasets for training. In addition, because the model is trained from end to end, there are no intermediate results for diagnosis purposes, and this exacerbates the black box behavior. The pipeline approach, on the other hand, is considered more tangible because it can output intermediate results for purposes of human inspection and validation (i.e., object detection bounding boxes). However, the pipeline approach is often sub-optimal because by training the sub-modules separately, it may lose track of the ultimate goal. Such segregated training can lead to a waste of computation resources due to the detection of irrelevant objects or the erroneous neglect of important objects. For example, the detection of objects located beyond the roadway sidewalk will not be beneficial to AVs. However, failure to detect traffic signal colors could be catastrophic. Another limitation of the pipeline approach is that it requires an explicit definition in the manner of cooperation of the two sub-modules; if the cooperation protocol is ill-defined, the overall performance of the entire model can be jeopardized even if the individual sub-modules exhibit good performance.

    Inspired by the work of Xu et al (*8*), the present study proposes a way to integrate the benefits from both settings by adding intermediate output heads and associating each head with corresponding loss while training the entire system end-to-end. For example, by adding "explanation head" to AV systems, the model can simultaneously output both driving actions and corresponding explanations. The explanations provide an opportunity for validating whether the correct causal relationships (between the driving environment from the input images and actions) have been learned. They also serve as extra labels to facilitate the entire training process. From the application perspective, the output explanations can help enhance the model interpretability, boost confidence in the model, and incentivize AV system manufacturers to adopt the technology.

## Imitation of human vision using image attention
A number of DL models have shown promising results in the recent years. However, human vision still remains as an incontrovertible benchmark. This is because the human eye possesses a multi-resolution structure, namely peripheral and foveal vision (*14*), and a proper "attention" mechanism to reach a balance in efficiency and recognition accuracy. Peripheral vision is blurry (low resolution) but requires only a short time for processing and has a larger field of vision. Foveal vision, on the other hand, is clear (high resolution) but requires longer processing time and only has a limited vision field. The combination of 2 structures guarantees the efficiency because it facilitates human to "attend" only to the salient and important regions with foveal vision while the overall information of the scene can be necessarily understood with only peripheral vision.

To imitate such human vision mechanism, visual attention has gradually evolved into a research topic of great interest to AV researchers. A state-of-the-art paper in end-to-end AV systems recently proposed an object-induced attention mechanism to generate driving decisions with "salient" objects in the scene (*8*). More recently, it has been demonstrated theoretically that a self-attention based model named Transformer (*15*) exhibits similar characteristics and performance as the convolutional neural network (CNN) but also possesses other ideal capabilities including the capture of long-range correlations within an image. The realization of such potential serves as the key inspiration and motivation of the current study. In subsequent sections of this paper, we demonstrate the efficacy of the Transformer based model in generating driving actions and explanations for autonomous driving systems.

**Research Gaps and Main Contribution of this Paper**
Xu et. al. proposed an object-induced attention mechanism (*8*) that performs an attention over the detected objects and uses only the relative objects to generate driving actions and explanations. More specifically, their model utilizes a "selector" structure to "crop" the fused regional features. This fused regional feature is generated by concatenating the regional features that are computed from FasterRCNN's region proposal network (*16*) (referred to as the local branch) and the raw overall feature map for the entire image (referred to as the global branch). The selector assigns a score to each region proposal to measure its relative importance to the overall driving task, and identifies the k regions with k-highest scores to compute the driving actions and explanations. Even though the model demonstrated satisfactory performance in predicting the actions together with explanations, there is still good reason to consider this attention mechanism as suboptimal and as having certain shortcomings.

The shortcomings arise due to two reasons. First, the ablation study results in (*8*) depict the baseline model with a "global" branch only, exhibits similar performance as the full model (which integrates the "global" and "local" features of the image). This indicates that the global features (overall information of the image) are relatively more important than the regional features (each object in the scenario). Also, this is consistent with the intuition that when generating high-level driving actions (move forward, turn left/right or stop), human drivers tend to use peripheral vision because only an approximate characterization of the driving scene is needed. For example, if there is an obstacle in the driving scene, the driver only needs to roughly see it (with peripheral vision) and can quickly eliminate the action of driving towards the obstacle's direction without clearly perceiving its shape or color. However, the object-induced attention as described in previous research (*8*) is more consistent to foveal vision because the model can "only" have a clear look into those cropped-out regions. Secondly, this attention mechanism largely depends on the object detection module (region proposal network), which has been pretrained to focus more on "object-containing" regions while ignoring "non-object" (background) regions. However, for driving tasks, as these "non-object" regions may contain vital information such as lane markers and drivable areas. Therefore, building the model on top of the method based on object detection, can be suboptimal.

To overcome these two shortcomings, we propose a global soft attention (GSA) mechanism which is similar to the peripheral vision capability of the human eye and uses the global features of the image. Intuitively, the model "softly" fuses the information from each region inside the image using Transformer model. The Transformer model is adopted here because multiple research studies have demonstrated that, compared to CNN, Transformer releases the constraints of generating visual features only based on local regions (*15*). This makes the model capable of possessing a "broader" horizon and capturing regions that not only are much wider compared to the traditional CNN kernel but also facilitate analysis of correlations within the image. This issue is further discussed in the results section of this paper.
In summary, the main contributions of this paper are threefold:
- Developed an end-to-end explainable DLCV model to generate driving actions with explanations.
- Proposed a new DL architecture with a novel visual attention mechanism using the Transformer model to achieve SOTA with significantly superior performance and lower computational cost compared to the benchmark model.
- Conducted multiple experiments that demonstrated the efficacy of the proposed attention mechanism in facilitating learning in a variety of settings.

## METHODS
As shown in **Figure 1**, the proposed model contains three blocks, namely, Feature Extractor, Multi-head Self Attention and Decision/Reason Generation.

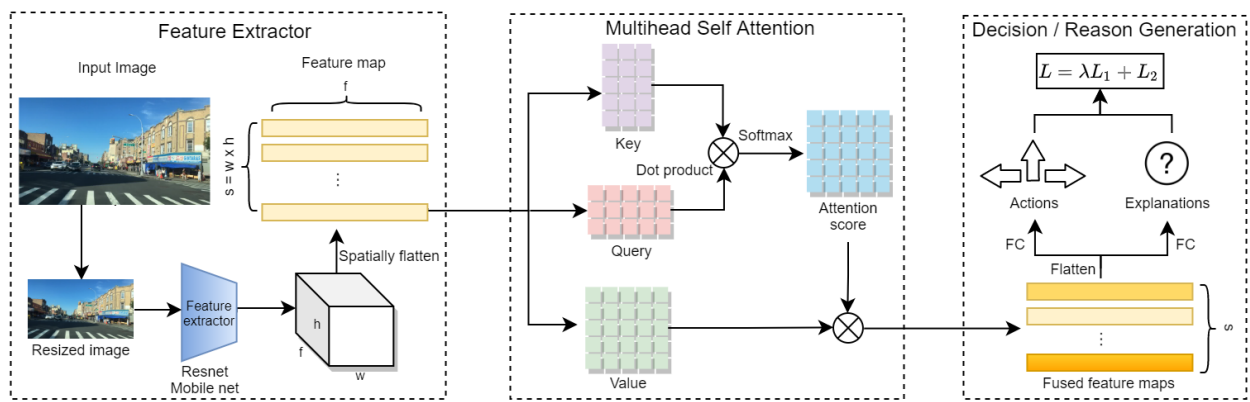

**Figure 1. Proposed model architecture**

### Feature Extractor
The Feature Extractor block first preprocesses the image (resize and normalization) and then computes the visual features using pre-trained Convolutional Neural Network (CNN) models. The output from the Feature Extractor block is a 3D tensor of shape $(h \times w \times f)$ where $f$ is the feature dimension of each spatial location and $h$, $w$ represent the height and width of the feature map. Then, the 2 spatial dimensions ($h$ and $w$) are flattened, and the output 2D feature map $X \in \mathbb{R}^{s \times f}$ is treated as a sequence of input with a sequence length of $s = h \times w$. In this work, we experimented with two backbone CNN models, namely, Resnet50 (*17*) or Mobilenet_v2 (*18*). Mobilenet_v2 has much higher computational speed due to its use of a smaller number of weights. On the other hand, the Resnet model is believed to represent the state of the art in image feature generation. The Feature Extractor module is frozen in both training and testing time since the main purpose for this research is to examine the performance of the upper stream architecture.

### Transformer
The feature map obtained from the Feature Extractor contains the information of the entire image, which is then fed to the Transformer to perform "global soft attention". The Transformer model is built with the basic block termed the multi-head self-attention (MHSA) layer. In this research, we adopt one MHSA layer to demonstrate its capacity. MHSA is a parallel computation of self-attention (SA) which utilizes the dot product to measure the "similarity" between the two inputs. More specifically, SA first computes three representations named query ($Q$), key ($K$) and value ($V$) with three separate linear layers: $Q = XW_Q$;

*Dong, Chen, Zong, Chen, Miralinaghi, Labi*

$K = XW_K$; and $V = X^T W_V$, where $W_Q, W_K, W_V$ are the corresponding weights, $K, Q \in \mathbb{R}^{s \times d_k}$ and, $V \in \mathbb{R}^{s \times d_v}$). Then a matrix of attention score $(a)$, which measures the correlation of the regions, is generated by conducting a dot product of the query and the key, followed by softmax normalization, as follows:

$$a = softmax\left(\frac{QK^T}{\sqrt{d_k}}\right) \in \mathbb{R}^{s \times s} \qquad (1)$$

Where $d_k$ is the dimension of K and Q. This attention mechanism enables the output embedding for each spatial location contains not only the information of the spatial location itself but also important information from other spatial locations. The attention scores serve as the fusion weights for generating the attended feature maps. The output of the SA is the multiplication of the value (V) and the attention score, and this completes the computation for single head.

$$head = Attention(K, Q, V) = aV \qquad (2)$$

Then the MHSA layer is simply the parallel version of SA which simultaneously computes multiple SA by concatenating all the heads, as follows:

$$MHSA(X) = concat(head_1, \ldots, head_h) W_{out} \in \mathbb{R}^{s \times d_{out}} \qquad (3)$$

Where $W_{out} \in \mathbb{R}^{d_v \times d_{out}}$ is the weights for the final output linear layer and $h$ is the number of heads. Compared to single head, using MHSA enables each head to simultaneously focus on different tasks and can attend to regions with different ranges. This manipulation can greatly enhance the model's flexibility and generalization power. The output from the MHSA layer has the same spatial dimension as the input feature map $X$. However, each spatial location contains the "fused" information from this location itself and other regions based on the automatically-computed correlation.

**Decision / Reason Generator**
The Decision/Reason Generator block is a standard multitask classifier containing two branches for generating driving actions and explanations, respectively. It takes the output feature map from MHSA block as input, and performs two separate classifications using fully connected (FC) layers. The model is trained end-to-end, following the classic multitask learning manner that aggregates the two losses (driving action loss $L_A$ and explanation loss $L_E$).

$$L_A = \sum_i^4 L(\widehat{A}_i, A_i); \quad L_E = \sum_i^{21} L(\widehat{E}_i, E_i) \qquad (4)$$

$$L = \lambda L_A + L_E \qquad (5)$$

Where $\lambda$ is the weight parameter for tuning the tradeoff between the two losses. Based on the experiment in (*8*), when $\lambda = 1$, the model yields the best performance in terms of both action and explanation prediction. In this work, we adopt this conclusion from that study, and use $\lambda = 1$. 4 and 21 in Eq.4 are the number of actions and explanations in the dataset, which will be explained in detail in the following Experiment Setting section. $L(\cdot,\cdot)$ is the binary cross entropy loss defined as follows:

$$L(y, \hat{y}) = -\frac{1}{N}\sum_{i=1}^N y_i \log(p(\hat{y}_i)) + (1 - y_i) \log(1 - p(\hat{y}_i)) \qquad (6)$$

## EXPERIMENTAL SETTINGS

### Baseline Models and Setups

A key technical motivation of this study is to test different attention mechanisms. Therefore, we compare our global soft attention (GSA) model with several baseline models including: regional hard-attention (RHA), regional soft-attention (RSA), and global no-attention (GNA) model.

- The RHA model is similar to the object-induced attention model proposed in a benchmark study (*8*) albeit with the local branch only. This model utilizes Faster RCNN with Feature Pyramid Network (FPN) as feature extractor as shown in **Figure 2 (a)**. followed by a Regional Hard Attention module which utilizes a fully connected (FC) layer to compute a score for each region proposals and select only top-k objects for generating actions (as shown in **Figure 2 (b)**). In this research, we trained 2 models with k=5 and k=10. We keep the local branch only to test the importance of regional information for the overall driving decision and explanation generation.
- The RSA model uses the same soft attention mechanism (Transformer) as the proposed GSA model (Multihead Self Attention block in **Figure 1**), but the attention is conducted over the region proposals instead of the global features. It uses the same FasterRCNN (FPN) as (*8*) to generate the regional features. This model mainly serves to compare the performance between soft attention and hard attention. In addition, we trained two RSA models with 5 and 8 heads.
- The GNA model mainly serves as an ablation study to our GSA model. It uses the same global features as GSA (generated from Resnet backbone) but replaces the Transformer structure (MHSA) block by a vanilla fully connected network (FCN) with a number of parameter similar to that of the MHSA block.

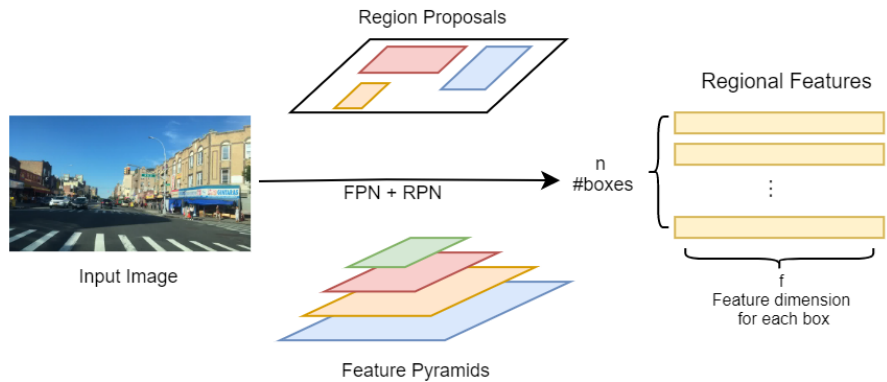

(a) Regional feature extractor

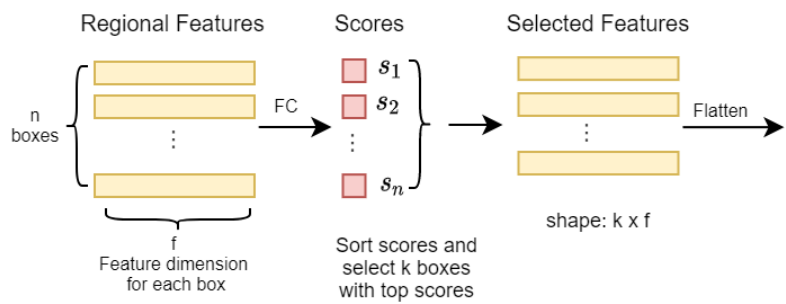

(b) Regional hard attention

**Figure 2. Two baseline structures**

**Dataset**
All the above-mentioned models were trained and evaluated using the BDD Object Induced Actions (BDD-OIA) dataset (*8*). This dataset extends the BDD-100K dataset (*19*) by labeling each frame individually with driving actions and explanations. The actions refer to high-level feasible driving maneuvers that can be undertaken by the driver at any specific time step: move forward, stop/slow down, turn left and turn right. The explanations are associated with the actions and are summarized into twenty-one (21) classes**. Figure 3** illustrates the example image and labels and. All the models are trained and **Table 1** presents the labels for actions and explanations. The model is developed with a training set of 16,082 images, a validation set of 2,270 images, and a test set of 4,572 images.

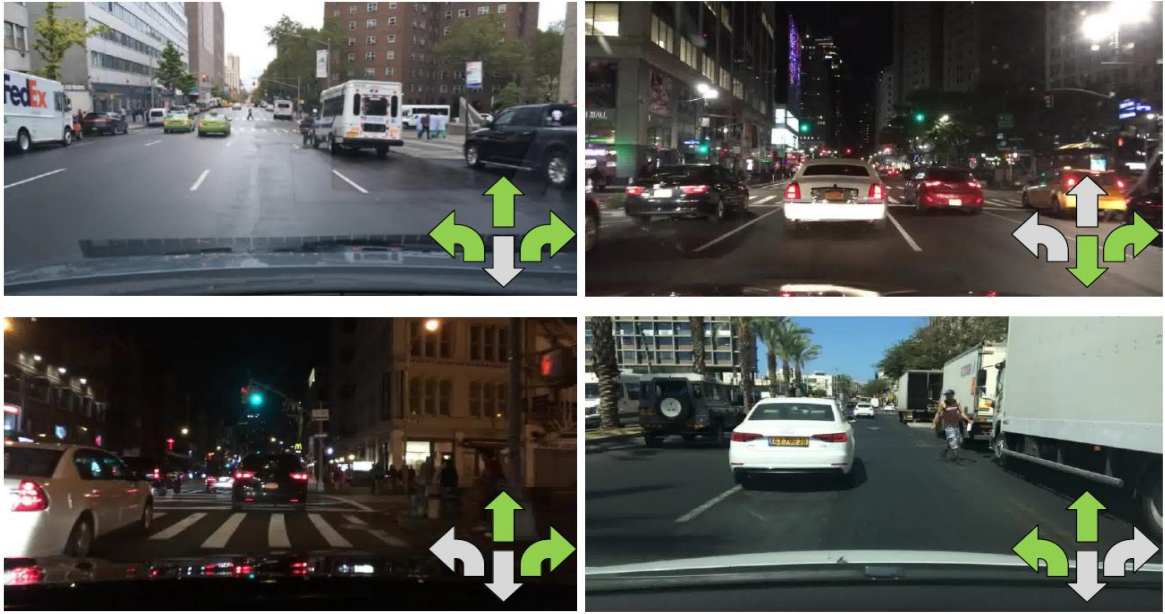

**Figure 3. Examples of groundtruth images and decisions from the BDD-OIA dataset (*8*)**

**TABLE 1 Actions With Explanations in BDD-OIA**

| Actions | Explanations |
|---|---|
| Move forward | Traffic light is green |
|  | Follow traffic |
|  | Road is clear |
| Stop/Slow down | Traffic light is red |
|  | Traffic sign |
|  | Obstacle: car |
|  | Obstacle: person |
|  | Obstacle: rider |
|  | Obstacle: others |
| Turn left | No lane on the left |
|  | Obstacles on the left lane |
|  | Solid line on the left |
|  | On the left-turn lane |
|  | Traffic light allows |
|  | Front car turning left |
| Turn right | No lane on the right |
|  | Obstacles on the right lane |
|  | Solid line on the right |
|  | On the right-turn lane |
|  | Traffic light allows |
|  | Front car turning right |

*Dong, Chen, Zong, Chen, Miralinaghi, Labi*## RESULTS
In this section, both training and testing results for the abovementioned experiments are documented.

**Training Curve**

All the models are trained using batch size $b = 10$, the stochastic gradient descent (SGD) method with an initial learning rate $\alpha = 0.001$, and a learning rate decay of $10^{-4}$. All the models are trained for 40 epochs (64,080 steps). **Figure 4** presents the corresponding training curves for our proposed GSA model and all the baselines mentioned above. From the training curve, it can be observed clearly that the two proposed GSA models converge much faster with much lower training loss compared to the baselines. The number of parameters and the resulting total training time varies across the models. **TABLE 2** presents the number of parameters and the training time for 40 epochs (trained on one NVIDIA Quadro RTX-6000 GPU, 24G RAM computer). The results indicate that even with fewer parameters and much lower training time, the proposed GSA model achieves lower training loss, and yields global attention that is generally superior to regional attention in terms of convergence rate and final training loss.

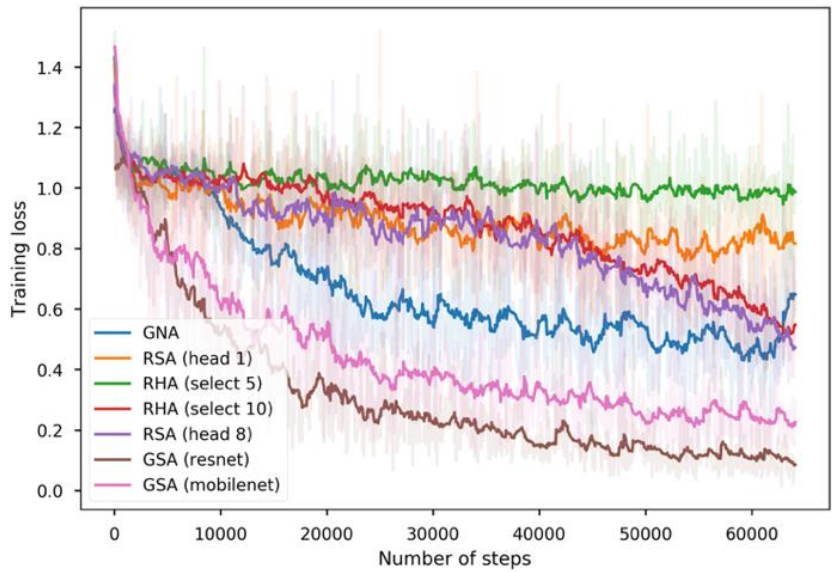

**Figure 4. Training curve for all 7 models**

**TABLE 2  Number of Parameters and Training Time**

| Model | # of parameters | Training time (40 epochs) |
|---|---|---|
| RHA (select 5) | 11.04M | 10h, 28min |
| RHA (select 10) | 21.53M | 10h, 30min |
| RSA (head 5) | 21.22M | 10h, 32min |
| RSA (head 8) | 20.75M | 10h, 42min |
| GNA | 26.10M | 3h, 10min |
| **GSA (resnet)** | **24.08M** | **3h, 15min** |
| **GSA (mobilenet)** | **2.61M** | **2h, 53min** |

**Comparative Results**
We performed the same prediction task (actions and explanations prediction) as the benchmark model presented in previous research (*8*). For evaluating our model, we used the same evaluation metrics used in the previous research. These metrics represent two versions of the F-1 score, namely, overall F1 score ($F1_{all}$) and mean in-class F1 score ($mF1$) for both decision and explanations. This is a common metric used in unbalanced data. A higher F1 score means the model has both higher precision and higher recall. More specifically, $F1_{all}$ is the F1 score calculated over all the predictions, and is calculated as follows:

$$F1_{all} = \frac{1}{|A|} \sum_{j=1}^{|A|} F1(\widehat{A_j}, A_j) \quad (7)$$

Where $\widehat{A_j}$ is the predicted value and $A_j$ is the true label (this may represent an action or an explanation), $|A|$ is the total number of predictions. Meanwhile, the dataset is unbalanced, in other words, there are more examples associated with the going-straight action compared to the turn-left action. Therefore, we calculated the F1 score for each predicted class, and then determined the $mF1$ value as the mean of all these F1 scores, as follows:

$$mF1 = \frac{1}{C} \sum_{j=1}^{C} \sum_{i=1}^{n} F1(\widehat{A_i^j}, A_i^j) \quad (8)$$

$C$ is the number of predicted classes (4 for actions, 21 for explanations), $n$ is the total number of points in the test dataset. **TABLE 3** presents the results for each model.
From **TABLE 3**, the following five (5) conclusions can be made:
1) The proposed global soft attention (GSA) models outperform all the baselines with a significant margin, particularly with regard to explanation prediction.
2) Global features are generally more useful compared to regional features, even where the vanilla model (GNA) is used without any special attention mechanism.
3) Soft attention is superior to hard attention even in cases where only regional information is available.
4) With regard to the feature extractor, using Mobilenet_v2 has comparable predictive performance compared to Resnet50 but saves a significant amount of training time.
5) Increasing the number of heads can generally enhance the performance of soft attention models.
These results also match the training loss curve as shown in **Figure 4**. **Figure 5** presents the example predictions (generated from GSA-Mobilenet) for four randomly chosen images in the test dataset.

**TABLE 3. F-1 Scores of Proposed Model and Baselines**

| Attention mechanism | Model | Decision mF1 | Decision $F1_{all}$ | Explanation mF1 | Explanation $F1_{all}$ |
|---|---|---|---|---|---|
| Regional Attention | Xu et, al.(*8*) | 0.718 | **0.734** | 0.208 | 0.422 |
| | RHA (select 5) | 0.572 | 0.494 | 0.482 | 0.047 |
| | RHA (select 10) | 0.565 | 0.495 | 0.499 | 0.123 |
| | RSA (head 5) | 0.595 | 0.476 | 0.506 | 0.127 |
| | RSA (head 8) | 0.608 | 0.542 | 0.554 | 0.330 |
| Global no attention | GNA (resnet) | 0.706 | 0.660 | 0.561 | 0.352 |
| **Global soft attention** | **GSA (resnet)** | **0.750** | **0.729** | **0.644** | **0.525** |
| | **GSA (mobilenet)** | **0.746** | **0.718** | **0.642** | **0.531** |

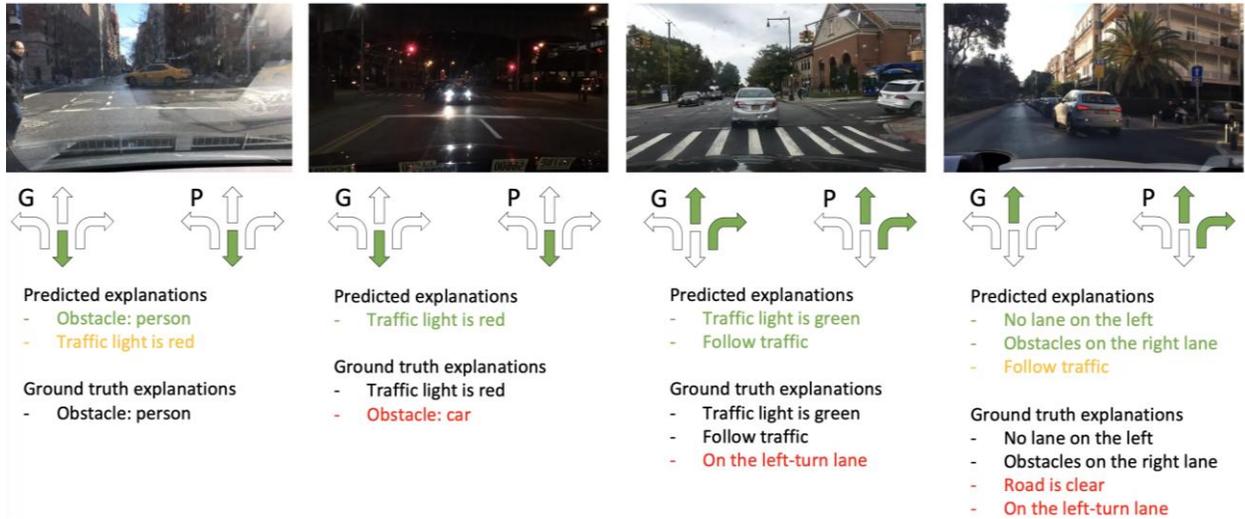

**Figure 5.** Example predictions. (Regarding action prediction, "G" stands for ground truth and "P" for model prediction. Regarding explanations, green color indicates true positive, yellow indicates false positive, and red indicates false negative)

**Analysis of the Results**
The observations expressed in the preceding paragraph can be attributed to the following reasons:
1) The global features are superior to regional features due to the inherent nature of driving decisions. For generating high-level actions (for example, move forward, stop/slow), the acquisition of an overall characterization of the roadway scene is more essential compared to the recognition of every single object and computation of their bounding boxes. Therefore, even the GNA baseline can yield superior performance compared to regional attention models built on top of object detection models.
2) The soft attention (Transformer in this work) is superior to the hard attention (score-based selection) because the former is capable of fusing individual pieces of information in the image based on their individual contributions to the ultimate driving goal (maneuver) rather than simply picking the more important regions. The latter inevitably creates a "bottleneck" to the information flow path and therefore leads to neglect of some information that could be useful to the driving maneuver.
3) The Transformer based models (the two GSAs) outperform the GNA because the MHSA structure they contain has the capability of capturing long-range correlations within an image. Compared to classic CNN based methods, which can only capture the local region correlations due to the fixed size of convolution kernels in each layer, the Transformer-based models enable information fusion over the entire image. This long-range correlation is typically crucial for driving decisions because there exists a "relativity" correlation within the image. For example, "left" is relative to the "right"; therefore, in order to generate the decision of "turn left", the model needs to understand which part of the image depicts the "left region". Since the cameras are not always facing the same direction as the movement direction of the vehicle, the ratio of "left region" to the entire image keeps changing. Therefore, the model has to understand "left" and "right" relatively from the scene context, which can be achieved with Transformer based model by simultaneously attending to multiple regions. This entire mechanism is analogous to the peripheral vision of the human eye as human drivers generating driving actions (quickly looking at multiple regions and then making driving decisions instantaneously without clearly seeing each individual object in the region) (*20*)(*21*).

## CONCLUSION
In this paper, we propose a novel architecture to generate driving actions as well as explanations based on images. The objective is to mitigate the low interpretability nature of deep learning based computer vision models and ultimately, to enhance user trust of autonomous driving systems. The proposed architecture utilizes the Transformer model (or, more specifically, the Multi-head Self Attention module) to imitate the peripheral vision of human in driving and therefore, to achieve SOTA on BDD-OIA dataset. The results from the experiments demonstrate that the proposed model outperforms all the baseline models in terms of prediction accuracy and training time. From the perspective of application, the proposed model can perform as a driving alarm system for both human-driven vehicles and autonomous vehicles because it possesses the capability to quickly understand/characterize the environment and to filter out infeasible driving actions. In addition, the extra explanation head of the proposed model provides an extra channel for sanity checks to guarantee that the model learns the ideal causal relationships. This provision is critical in the development of autonomous systems.

Moving forward, the proposed model can be further improved by incorporating and fusing other sources of information such as LiDAR point clouds and information from vehicle-to-vehicle (V2V) connectivity (*22–29*) . The fused information offers additional safety benefits by providing the autonomous driving system with redundancy, that is, in case a sensor misfunctions, the system can still operate with information from other sensors.

## ACKNOWLEDGMENTS
This work was supported by Purdue University's Center for Connected and Automated Transportation (CCAT), a part of the larger CCAT consortium, a USDOT Region 5 University Transportation Center funded by the U.S. Department of Transportation, Award #69A3551747105. The contents of this paper reflect the views of the authors, who are responsible for the facts and the accuracy of the data presented herein, and do not necessarily reflect the official views or policies of the sponsoring organization.
This manuscript is herein submitted for PRESENTATION ONLY at the 2022 Annual Meeting of the Transportation Research Board.

## AUTHOR CONTRIBUTIONS
The authors confirm contribution to the paper as follows: all authors contributed to all sections. All authors reviewed the results and approved the final version of the manuscript.